%% file: acl_latex.tex
\documentclass[11pt]{article}

\usepackage[final]{acl}

\usepackage{times}
\usepackage{latexsym}

\usepackage[T1]{fontenc}

\usepackage{wrapfig}

\usepackage[utf8]{inputenc} 
\usepackage[T1]{fontenc}    
\usepackage{hyperref}       
\usepackage{url}            
\usepackage{booktabs}       
\usepackage{amsfonts}       
\usepackage{nicefrac}       
\usepackage{microtype}      
\usepackage{xcolor}         
\usepackage{lipsum}
\usepackage{times}
\usepackage{latexsym}
\usepackage{amsmath}
\usepackage[T1]{fontenc}
\usepackage{color}
\usepackage{amsfonts} 
\usepackage{amssymb} 
\usepackage{colortbl}
\arrayrulecolor{black} 
\usepackage{hhline}
\usepackage{algorithm}
\usepackage{algorithmic}
\usepackage{multirow}
\usepackage{xcolor}
\usepackage{booktabs}
\usepackage{algorithm}
\usepackage{algorithmic}
\usepackage{xspace}
\usepackage{inconsolata}
\usepackage{enumitem}
\usepackage{rotating, multirow}
\usepackage{graphicx}
\usepackage{txfonts}

\title{ReFreeKV: Towards \textit{Threshold-Free} KV Cache Compression}

\newcommand\method{ReFreeKV\xspace}

\author{
Xuanfan Ni\thanks{~Equal contribution. Partial work done during Xuanfan's internship at Tencent. Correspondence to: Liyan Xu, Piji Li.}$^\clubsuit$~~~Liyan Xu\footnotemark[1]\footnotemark[2]$^\varheartsuit$~~~~Chenyang Lyu$^\vardiamondsuit$~~~Longyue Wang$^\vardiamondsuit$\\ 
\bf Mo Yu$^\varheartsuit$~~~\bf Lemao Liu$^\blacktriangle$~~~\bf Fandong Meng$^\varheartsuit$~~~\bf Jie Zhou$^\varheartsuit$~~~\bf Piji Li$^\spadesuit$ \\
$^\clubsuit$Nanjing University of Aeronautics and Astronautics\\
$^\varheartsuit$WeChat AI, Tencent~~~~$^\blacktriangle$Fudan University~~~~$^\vardiamondsuit$Independent Researcher\\
\fontsize{10pt}{\baselineskip}{\texttt{xuanfanni@nuaa.edu.cn~~~liyanlxu@tencent.com}}}

\begin{document}
\maketitle
\footnotetext[2]{~Project lead: Liyan Xu~\textless{\href{mailto:liyanlxu@tencent.com}{liyanlxu@tencent.com}}\textgreater.}

\begin{abstract}
To reduce memory consumption during LLM inference, a handful of methods have been proposed for KV cache pruning.
While these techniques can accomplish lossless memory reduction on many datasets, they often hinge on an under-emphasized condition: an input/domain-specific threshold for KV cache budget needs to be pre-determined to achieve the optimal performance.
However, such input-sensitive design may be considerably limited in real-world scenarios, as open-domain inputs span diverse domains, lengths and difficulty levels, without clear boundaries for threshold selection.
As a result, the dependence of such input-sensitive threshold can be a fundamental limitation that causes large degradation on arbitrary inputs.

In this work, we propose a new objective that lifts the threshold constraints for robust KV compression, advocating for ``\emph{threshold-free}'' methods that adaptively adjust budget allocation while preserving full-cache performance.
We then propose a novel method, ReFreeKV, serving as the first instantiation of this objective. Extensive experiments across 13 datasets with diverse context lengths, task types, and model sizes demonstrate its efficacy and efficiency. Our code is publicly released at \url{https://github.com/Patrick-Ni/ReFreeKV}.
\end{abstract}

\input{sections/intro}
\input{sections/related}

\input{sections/method}
\input{sections/exp}

\input{sections/conclusion}

\section*{Limitations}
\label{limitation}
The primary limitation of \method lies in the gap between its achieved compression ratio and the true optimal budget. As shown in Table~\ref{main_result}, in certain scenarios (e.g., QMSum with Mistral-7B), \method retains an 84.3\% budget, whereas a 50\% budget remains viable without performance degradation. We view this gap as an opportunity for future work to enable more aggressive KV-cache compression while still satisfying the full-cache objective under dynamic budgets.

Another limitation is that, although \method demonstrates near-lossless compression empirically, it does not provide formal guarantees on performance degradation. As reported in Table~\ref{main_result}, while \method even surpasses full-cache performance for Llama3-8B and Qwen2.5-7B, it incurs a small degradation (1.5\%) on Mistral-7B. Developing more principled approaches to further improve robustness remains an important direction for future research.

\bibliography{main}
\clearpage
\appendix

\input{apdx/freeze_old}
\input{apdx/datasets}
\input{apdx/ablation}
\input{apdx/slmf}
\end{document}

%% file: sections/intro.tex
\section{Introduction}
\label{sec:intro}

Transformer-based large language models (LLMs) generate text autoregressively and maintain a \textit{KV Cache} to store intermediate states during inference \cite{DBLP:journals/corr/abs-2307-03109,DBLP:journals/corr/abs-2303-08774,DBLP:journals/corr/abs-2402-06196}. At each decoding step, the model retrieves the cached key and value vectors of all previous tokens, which typically reside in GPU memory for attention computation \cite{DBLP:conf/nips/VaswaniSPUJGKP17,DBLP:journals/corr/abs-1911-02150,DBLP:conf/emnlp/AinslieLJZLS23}.
Consequently, managing the KV cache efficiently has become crucial to mitigate the overall memory consumption and inference overhead, as they grow proportionally with the model size and the input length.
For instance, Llama3-70B~\cite{DBLP:journals/corr/abs-2407-21783} demands a gigantic memory up to 50GB for 20K tokens.

Towards the KV cache efficiency, numerous recent methods have been proposed to effectively reduce the KV footprint after LLM prefilling. 
Exploiting the \emph{sparsity} of attention, prior works have demonstrated that retaining full KV cache is not always necessary.
Several optimization methods, such as H2O \cite{DBLP:conf/nips/Zhang00CZC0TRBW23}, ScissorHands \cite{DBLP:conf/nips/LiuDLWXXKS23}, SnapKV \cite{DBLP:conf/nips/LiHYVLYCLC24}, FastGen \cite{DBLP:conf/iclr/Ge0LZ0024}, CAKE \cite{DBLP:conf/iclr/QinCLHFCLL25}, etc., discard the less critical cache positions according to their designed pruning criteria. 
Other paradigms such as KVMerger~\cite{DBLP:journals/corr/abs-2407-08454} and D2O~\cite{DBLP:journals/corr/abs-2406-13035} resort to merge or compress KV vectors instead of hard-pruning for achieving the memory reduction effects.

However, almost all prior KV reduction techniques hinge on a fundamental yet often under-emphasized condition: a \textbf{data-dependent budget threshold} is typically involved to selectively tune for satisfactory results. For instance, D2O concludes a pre-defined KV cache budget ratio of 20\% to match full-cache performance on LongBench \cite{DBLP:conf/acl/BaiLZL0HDLZHDTL24}, whereas our experiments suggest that the required budget can rise to 80\% on GSM8K to maintain full performance. Similarly, retaining 1024 cache positions is sufficient for CAKE on LongBench, yet the same budget underperforms on the needle-in-the-haystack test \cite{needle}.

The existence of such threshold serves fine in idealized \emph{research settings} given dedicated datasets. Yet, its applicability may be considerably limited in \emph{real-world scenarios}, where the \textbf{\emph{inputs are intermixed across different domains, lengths and difficulty levels without explicit separation}}. As a result, optimal thresholds cannot be pre-determined for diverse real-world inputs, making the system less robust and prone to significant performance degradation in practice.

To further illustrate, Table~\ref{tab:intro} presents preliminary experiments using H2O, StreamingLLM~\cite{DBLP:conf/iclr/XiaoTCHL24} and SnapKV, where inputs from different datasets are mixed to demonstrate the drawback of data-specific thresholds: \emph{a threshold reaching full performance on one dataset may not transfer well to others}. As the KV cache budget ratio changes, their performance varies substantially across datasets.

\input{tab/intro}

Such inconsistency on the performance motivates us to revisit the goal of KV cache pruning. Rather than targeting only how much memory can be saved on a fixed benchmark, we instead ask for a \textbf{new objective}:
\emph{lifting the dependency of data-specific thresholds in KV cache pruning}, such that the system should robustly handle arbitrary inputs with \emph{consistent} full-cache performance.

Specifically, our objective prioritizes two principles: \textbf{1)} the method may operate with a universal threshold insensitive to inputs, effectively rendering it ``\emph{threshold-free}''; and \textbf{2)} the method should consistently achieve performance comparable to its full-cache counterpart, able to dynamically adjust KV cache budgets.
Pursuing the best possible compression ratio only comes after satisfying these two criteria. As shown by our full experimental results in Table~\ref{main_result}, while prior methods may achieve strong compression on certain datasets, none could fulfill our objective to achieve consistent performance across diverse inputs.

Towards this objective, we propose \textbf{\method}, a novel method featuring th\underline{Re}shold-\underline{Free} KV cache pruning.
\method adopts a two-stage process with an input-insensitive threshold metric to dynamically control the KV cache budget.
Conceptually, it first ranks all KV cache positions based on their positional importance; then, it progressively retains key-value vectors in order, and discards the remaining KV cache once the stopping criterion is met.
For minimal overhead, \method is designed for implementation with parallel operators, instead of sequential processing. As shown in Section~\ref{ssec:time}, its latency is on par with prior efficient KV pruning methods across varying batch sizes.

The \emph{threshold-free} aspect stems from a universal metric in the stopping criterion, termed \emph{Uni-Metric}, whose design is shown to be consistent and \textbf{insensitive to variations} in input domains and sequence lengths. 
It should be noted that \emph{threshold-free} does not mean the absence of any thresholding mechanism; though, we intentionally use this framing to emphasize that the efficacy of such method does not rely on a pre-determined threshold, thus no threshold calibration is needed in practice.
By design, \method naturally adjusts to higher compression ratios on simpler tasks, while allocating more cache resources to more complex ones.

\begin{figure*}[!t]
   \centering
   \includegraphics[width=0.8\textwidth]{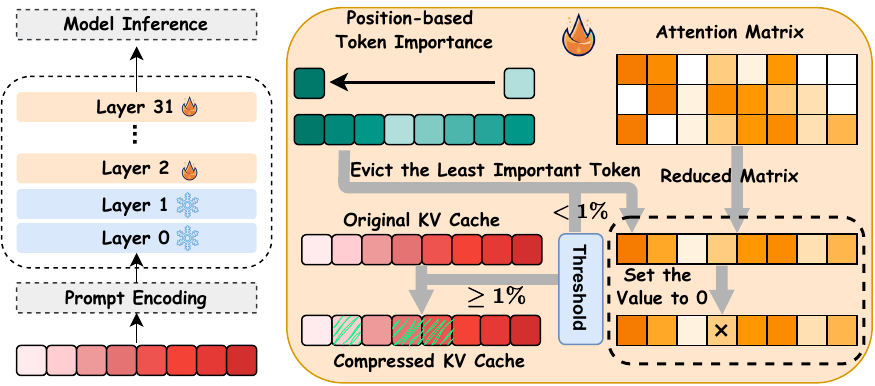}
        \caption{The overall workflow of \method in Section~\ref{ssec:method}. After prefilling, tokens are initially ranked based on their positions, followed by the eviction of the least significant tokens (per layer), whose halting condition is determined by the norm value of the reduced attention matrix. The KV cache for the remaining tokens are then preserved to subsequent generation.}
    \label{framework}
\vspace{-0.5em}
\end{figure*}

To evaluate \method, our experiments adopt 13 datasets varying diverse context lengths and tasks, including mathematical and commonsense reasoning, reading comprehension and coding, which show that \method accomplishes our proposed objective.
The resulting inference is highly comparable and can even surpass the full-cache performance, evaluated with multiple LLMs of different sizes, all without an input-specific threshold.
For instance, using Llama3-8B, it automatically allocates an average KV budget ratio of 63.7\%, while slightly exceeding full-cache performance across 13 datasets on average.
Though the best compression ratios on a few datasets are not achieved by \method, however, it remains the only method that could maintain decent compression under the \emph{consistent performance constraint}, successfully addressing the limitation of existing baselines requiring data-specific budget thresholds.

%% file: tab/intro.tex
\begin{table}[!t]
\centering
\resizebox{0.88\columnwidth}{!}{
\begin{tabular}{l|ccc|c}
\toprule[1.3pt]
& \textbf{GSM8K} & \textbf{GPQA} & \textbf{CoQA} &\textbf{Avg.} \\ \cmidrule{1-5} 
\multicolumn{5}{c}{\textit{Budget = 50\%}}                       \\
\cmidrule{1-5}
H20 & 41.29\% &	9.99\%	& 99.30\% & 50.19\%\\
SLM &	4.53\% &	20.78\%	&84.72\% & 36.68\%\\
SnapKV &17.22\% &	26.91\%	& 100.19\% &48.11\% \\
\cmidrule{1-5}
\multicolumn{5}{c}{\textit{Budget = 20\%}}                       \\
\cmidrule{1-5}
H2O &	3.23\%	& 8.48\% &	96.53\%	& 36.08\% \\
SLM &	4.26\%	& 11.54\%&	76.05\%	&30.62\% \\
SnapKV&	1.71\%	&3.86\%	&98.60\%	&34.72\% \\
\bottomrule[1.3pt]
\end{tabular}
}
\caption{KV pruning methods that depend on a preset KV budget threshold can exhibit inconsistent performance across domains (percentage relative to full-cache scores using Llama3), making input-specific threshold selection unavoidable for achieving optimal inference.}
\label{tab:intro}
\vspace{-1em}
\end{table}

%% file: sections/related.tex
\section{Related Work}
\label{sec:related}

\paragraph{KV Cache Pruning}
To mitigate the large memory footprint of the KV cache during LLM inference, a prominent line of work has focused on pruning, which selectively discards less important cache positions. Methods like Scissorhands~\cite{DBLP:conf/nips/LiuDLWXXKS23} and SnapKV~\cite{DBLP:conf/nips/LiHYVLYCLC24} retain tokens based on high attention scores. Others employ strategies based on token recency and historical importance, such as H2O~\cite{DBLP:conf/nips/Zhang00CZC0TRBW23} or StreamingLLM~\cite{DBLP:conf/iclr/XiaoTCHL24}. FastGen~\cite{DBLP:conf/iclr/Ge0LZ0024} further refines this by adapting retention strategies on a per-head basis.

\paragraph{Input-Sensitive Design in Recent Works}
While the shift from fixed-budget constraints to heuristic-based halting conditions is an established paradigm in recent literature, existing approaches often remain constrained by input-sensitive hyperparameters. Lethe \cite{DBLP:conf/aaai/ZengZYHZLJZ26} introduces layer- and time-adaptive pruning during decoding, but still relies on \textit{sparse} and \textit{recent} ratio without particular emphasis on cross-task stability. SABlock \cite{DBLP:journals/corr/abs-2510-22556} employs semantic-aware token scoring with adaptive block sizes, yet requires a pre-defined cache budget tuned per task. DuoAttention \cite{DBLP:conf/iclr/XiaoTZGYTF025} classifies heads into retrieval and streaming types with a constant-length cache for streaming heads, but this fixed window size is not designed for varying input complexity. Ada-KV \cite{DBLP:journals/corr/abs-2407-11550} proposes head-wise adaptive budget allocation guided by a theoretical loss bound, yet it essentially redistributes a pre-defined global budget across heads rather than eliminating the budget constraint itself. Although Twilight \cite{DBLP:journals/corr/abs-2502-02770} removes the explicit budget via a top-$p$ inspired mechanism, it shifts the burden to tuning the $p$ value across models and inputs. In contrast, ReFreeKV aims to eliminates task-dependent threshold search; through validation across diverse datasets and models, it preserves full-cache performance robustly without any manual tuning.

%% file: sections/method.tex
\section{Methodology}
\label{method}

In this section, we first elaborate our motivation and the new objective for KV cache compression differing from prior works.
We then delineate our proposed approach \method, along with its key implementation details.

\subsection{\emph{Threshold-Free}: A New Objective}
\label{ssec:obj}

As discussed in Section~\ref{sec:related}, prior methods of KV cache compression require a pre-selected budget threshold in various forms. While these methods can perform well on numerous datasets, it is inevitable that such dependence on a threshold can become a practical limitation.
As the optimal threshold can vary across different inputs, it is not always feasible for such systems to pick appropriate thresholds in maintaining stable performance involving arbitrary inputs and open-domain instructions.
Certain inputs may necessitate a relatively higher memory budget, such as tasks with multi-step or mathematical reasoning, whereas others such as straightforward QA queries need only a small set of KV cache. 
Inputs in real-world scenarios, especially, are intermixed and unpredictable, with no clear boundary by difficulty or domain.

Our propose objective is exactly to remove input-specific threshold constraints, motivating \emph{threshold-free} methods that ensure consistent performance comparable to full-cache regardless of inputs, while obviating the need of tuning for optimal thresholds. 
Pursuit of the best compression ratio is prioritized only after satisfying this aspect. To the best of our knowledge, we are the first to propose an effective solution that fulfills such objective.

\subsection{\method}
\label{ssec:method}

\method consists of two stages implemented with efficient parallel operators. Conceptually, it first ranks all KV cache positions per layer and per attention head; then, it sequentially retains key–value vectors until a stopping condition, determined by our input-insensitive threshold metric, is met.
The KV cache at those remaining positions is subsequently discarded.

\paragraph{The \emph{Two-Stage} Logistics}
The rationale behind a two-stage process is that precisely determining the optimal pruning positions in a sequence (i.e., subset selection) is inherently \emph{combinatorial}. By adopting an approximate ranking stage (as an inductive bias) followed by a one-time search, the problem becomes tractable. The entire procedure is further efficiently optimized through parallel operations.

In line with most prior KV compression works \cite{kv_survey}, our method is applied only once after input prefilling, with the primary goal of reducing memory consumption during inference and bringing improved throughput (Section~\ref{ssec:time}).

\paragraph{Initial Ranking}
The first stage ranks all KV cache, such that the beginning of the sequence may likely contain more critical information than its latter parts, which forms the basis for downstream sequential eviction.

At this initial stage, we exploit the properties observed by prior works.
First, positions at the beginning of the input generally play a more critical role in subsequent generation, known as \emph{attention sinks} \cite{DBLP:conf/iclr/XiaoTCHL24,sun2026spikesparsesinkanatomy}. 
Second, latest positions usually receive a greater attention ratio \cite{DBLP:conf/iclr/GuPDLZD0L25}.
Building on the \emph{positional bias} reported in previous works, we rank KV cache by token positions as follows.

Denote a LLM input sequence with $n$ tokens as $X=\{x_1,x_2,\dots,x_n\}$, where each Transformers layer originally consists of $n$ positions of KV vectors per attention head. The initial ranking takes the first $m$ positions and reversely takes the remaining $n-m$ positions, denoted by $\widehat{X} = \{x_1,x_2,\dots,x_m, x_n,x_{n-1},\dots,x_{m+1}\}$. $m$ is a chosen hyperparameter that works well regardless of specific input sequences.

Despite its simplicity, the position-based ranking is shown not only effective but also particularly advantageous in terms of computational overhead, comparing to other ranking strategies we conducted in Section~\ref{ssec:ablation}, therefore constituting the first stage of \method.
However, relying solely by positions does not fulfill our objective, as our experiments in Appendix~\ref{app:slmf} show that prior such methods such as StreamingLLM \cite{DBLP:conf/iclr/XiaoTCHL24} exhibit inconsistent performance across inputs, which highlights the need for a more robust KV eviction strategy.

\paragraph{Eviction by \textit{Uni-Metric}}
With the initial ranking on KV cache, \method then sequentially retains KV vectors, and halts upon the stopping condition by an input-insensitive threshold metric, termed \textit{Uni-Metric}, after which the remaining cache is effectively evicted.

The design of \textit{Uni-Metric} is then at the core of this process, which requires to signal the degradation level after removing KV cache of certain positions.
we propose a metric that empirically correlates well with the performance change when discarding a position, which could \textbf{serve as a bridge to ensure a minimal degradation upon the full cache performance}.
Inspired by~\citet{DBLP:journals/corr/abs-2406-11430}, we utilize the Frobenius norm (L2 norm) of the attention matrix $A \in \mathbb{R}^{n\times n}$ as the \textit{Uni-Metric}, denoted as $||A||_F = \sqrt{\sum_{i=1}^n\sum_{j=1}^n|A_{i,j}|^2}$.
For each position $i$ in the initially ranked sequence, we compare the Frobenius norm of the original attention matrix, $||A||_F$, with that of a curated attention matrix, $||\widetilde{A_i}||_F$, in which scores to all positions $>i$ in the ranked sequence are masked out, replicating the effect of discarding all KV cache beyond position $i$. Once the norm difference reaches a threshold $T$ at position $i_{\text{prune}}$, the entire process terminates, retaining only the KV cache up to $i_{\text{prune}}$ and discarding the remainder, denoted as:
\begin{align}
\label{eq:prune}
    i_{\text{prune}} = \text{argmin}_{j=1}^n (1 - \frac{||\widetilde{A_j}||_F}{||A||_F} < T)
\end{align}

\begin{figure}[!t]
\centering
\includegraphics[width=0.85\columnwidth]{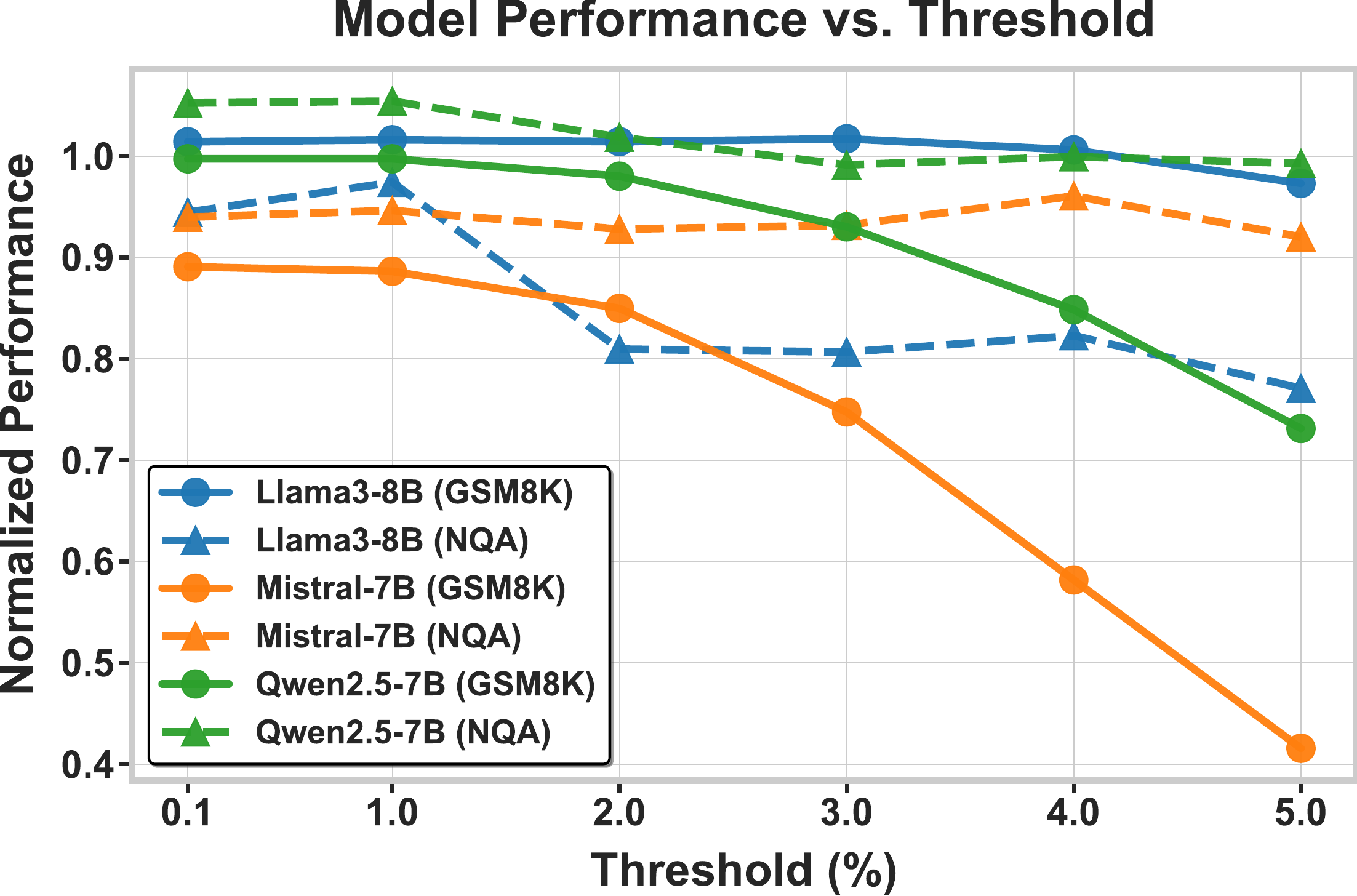}
\caption{Performance trends of Llama3-8B, Mistral-7B, and Qwen2.5-7B across varying \textit{Uni-Metric} thresholds. The x-axis represents the threshold percentage (0.1\% to 5\%), and the y-axis denotes the performance score normalized by the full-cache performance.}
\label{fig:method}
\end{figure}

\paragraph{The Universal Threshold}
To fulfill our objective, the threshold $T$ should ensure near lossless pruning invariant to inputs. Upon empirical search, we identify $T=1\%$ could serve well for this purpose.
Figure~\ref{fig:method} illustrates how performance varies with changes in the norm across different domains. Preliminary studies indicate that when $T<1\%$, performance remains comparable to the full-cache version robustly. We select $T=1\%$ to balance minimal degradation with maximal cache eviction. The efficacy of \textit{Uni-Metric} and its universal threshold is validated at full scale in the main experiments presented in Section~\ref{ssec:result} and Figure~\ref{fig:combined_tradeoff}.

\paragraph{Reducing Overhead}
As the input sequence length $n$ increases, the time and space overhead for norm calculation on the attention matrix grows by $O(n^2)$. To reduce the computational scale, we seek to use an approximate norm calculation by $O(n)$. 
Instead of using the full attention $A$, we reduce $A$ by taking the average of its last $k$ rows to a single attention vector $A'\in\mathbb{R}^{1\times n}$. The score for a position $i \in [1,n]$ in $A'$ is denoted as:
\begin{align}
A'[i] = \frac{\sum_{j=k}^{n}A_{i,j}}{\sum_{j=k}^n\mathbf{1}_{\{A_{i,j}\neq 0\}}}
\end{align}

\subsection{Implementation Details}
\label{ssec:impl}

As \method is conceptually a sequential search process, the design of \method allows efficient implementation by PyTorch's operators, such that the stopping positions of all layers are directly identified in parallel without explicit looping operations. Latency and throughput analyses in Section~\ref{ssec:time} demonstrate that \method matches prior popular KV pruning approaches, with negligible latency overhead and improved throughput compared to the baselines.

Specifically, the pruning position $i_{\text{prune}}$ can be determined directly by the combination of Torch \texttt{cumulative-sum} and \texttt{where} operators. Given the reduced attention matrix after the initial ranking, $A'_{\text{rank}}$, we compute the cumulative square-sum of each element, such that $A'_{cumsum}[i]$ represents the Frobenius Norm of the attention matrix after removing all cache to the right of position $i$ in $A'_{\text{rank}}$:
\begin{align}
||\widetilde{A'_i}||_F = A'_{cumsum}[i] = \sqrt{\sum_{k=1}^{i} \Big( A'_{\text{rank}}[k] \Big)^2}
\end{align}
We then divide $||\widetilde{A'_i}||_F$ by the full norm of $A'_{\text{rank}}$ to determine the norm difference as in Eq~\eqref{eq:prune}. The torch \texttt{where} operation allows us to directly identify the leftmost position that satisfies the $1\%$ universal threshold, ultimately yielding the set of positions for which the KV cache is called to retain. 
The overall pruning process of \method is further presented in Algorithm~\ref{al}.

\paragraph{Retaining Bottom Layers} Aligned with prior works \cite{DBLP:journals/corr/abs-2406-13035,DBLP:conf/iclr/XiaoTCHL24}, we identify that the LLM's initial layers have a relatively uniform attention distribution, usually requiring to retain most of the cache positions, as early layers are important for semantic understanding \cite{NEURIPS2019_159c1ffe,skean2025layer}. 
For simplicity and robustness, we always retain full KV cache of the first two layers in our implementation. We provide more studies on retaining KV cache of bottom layers in Appendix~\ref{app:freeze}.

\paragraph{Running at Scale} Supporting batch sizes $>1$, \method is able to perform the entire pruning process for each sample in parallel. To achieve this, we pad the shorter cache segments and update the attention masks accordingly, allowing the LLM to ignore the padded KV positions. The padding operation has a negligible impact on overall performance. Meanwhile, in popular LLM inference engines, e.g. vLLM~\cite{kwon2023efficient}, it is possible to allocate separate KV cache size for each sample, which aligns well with \method.

%% file: sections/exp.tex
\section{Experiments}
\label{sec:exp}
\input{tab/main.tex}
\subsection{Experimental Settings}
\label{ssec:ex-set}

\paragraph{LLM Backbones} Our experiments are conducted with three LLM families of different model sizes: Llama3-Instruct with size of 8B/70B, Mistral-7B-Instruct-V0.3, and Qwen2.5-Instruct with size of 7B/32B/72B. We implement our \method upon the released codebase of SnapKV\footnote{~\url{https://github.com/FasterDecoding/SnapKV}}. For the reduced attention matrix $A'$, we set $k=1$ in practice (ablation provided in Section~\ref{ssec:ablation}), and set $m=4$ for the initial ranking stage.

\paragraph{Datasets} For comprehensive evaluation, we evaluate \method on datasets of both short and long context length of different domains, including mathematics, science, and commonsense reasoning on GSM8K \cite{DBLP:journals/corr/abs-2110-14168}, GPQA \cite{DBLP:journals/corr/abs-2311-12022}, TheoremQA \cite{DBLP:conf/emnlp/ChenYKLWMXWX23}, TruthfulQA \cite{DBLP:conf/acl/LinHE22}, and CoQA \cite{DBLP:journals/tacl/ReddyCM19}. We also include tasks from Longbench \cite{DBLP:conf/acl/BaiLZL0HDLZHDTL24} with 8 datasets spanning document comprehension, summarization and coding.
Appendix~\ref{app:datasets} provides a detailed description and statistics for all \textbf{13 datasets}, along with how they are utilized in our experiments.

\paragraph{Evaluation Protocol} 
Our evaluation primarily assesses \method’s ability to preserve full-cache performance with its automatic pruning budgets. Accordingly, we compare \method against its full-cache scores, and also report the average compression ratio for each dataset.

Additionally, we compare with five prior KV cache pruning methods with varied budget sizes, including: Heavy Hitter Oracle (\textbf{H2O}) \cite{DBLP:conf/nips/Zhang00CZC0TRBW23}, StreamingLLM (\textbf{SLM}) \cite{DBLP:conf/iclr/XiaoTCHL24}, \textbf{SnapKV} \cite{DBLP:conf/nips/LiHYVLYCLC24}, \textbf{PyramidKV}~\cite{DBLP:journals/corr/abs-2406-02069} and \textbf{CAKE}~\cite{DBLP:conf/iclr/QinCLHFCLL25}.
By evaluating performance consistency under fixed KV cache budgets of 90\%, 50\%, and 20\%, we further illustrate the limitations arising from the input-specific threshold dependence.

Lastly, we also include a concurrent work \textbf{Twilight} \cite{DBLP:journals/corr/abs-2502-02770}, which does not rely on a budget threshold but employs a top-$p$-inspired metric for adaptive token selection. It is worth noting that the $p$ value remains a hyperparameter to be tuned across models and inputs.

\subsection{Main Results}
\label{ssec:result}

The main experimental results are shown in Table~\ref{main_result}.
For comparison with Twilight, we separately report the results in Table~\ref{tab:twi} due to its different hyperparameter type. Based on Table~\ref{main_result}, we can draw the following observations.

• \method is able to fulfill our objective, capable of performing near-lossless dynamic compression across different models, varying input lengths, and diverse task types. Interestingly, with Llama3-8B and Qwen2.5-7B, \method even surpasses the full-cache performance by 0.12\% and 2.63\% respectively, utilizing an average of 63.68\% and 76.02\% KV cache. With Mistral, \method also manages to achieve near 15\% compression with a relatively small 1.5\% performance reduction. These results indicate that our proposed method can strike for real-world scenarios with no bother by input-specific budget thresholds. 

• In stark contrast, previous methods achieve a consistent full-cache performance only when manually determined a high budget ratio, e.g. 90\%. However, when the budget is reduced, e.g. 50\% and 20\%, the degradation can become severe on certain datasets, distinct from \method that automatically adjusts the pruning to always prioritize full-cache performance. Theoretically, one could tune the budget for each dataset that achieves minimal degradation, but this is generally infeasible for real-world open-domain instructions.

• \method naturally reflects the \emph{difficulty} of the generation task. As in Table~\ref{main_result}, the dynamic budget ratio is high on Math\&Science datasets (over 90\%), while much lower on QA or Summarization datasets (as low as 15\%). This observation is in line with our intuition, where inference on concise but hard tasks, such as math problems, requires more context and more precise calculation, resulting in higher budget allocation. From this aspect, our method design serves beyond for memory efficiency, but could be potentially leveraged for input analysis in a broader scope.

• Besides the dynamic compression, \method also outperforms the three 90\%-budget baselines, while itself uses less than 90\% budget. 
Though, It is worth reiterating that the goal of this work is not to propose yet another KV cache pruning method that targets the best possible compression ratio under specific conditions.
Instead, we seek to lift the threshold constraints and advocate for robust KV pruning that generalizes to arbitrary inputs.

\input{tab/twi}

\input{tab/time}

As separately reported in Table~\ref{tab:twi}, we compare \method with Twilight's reported performance across the GSM8K, NarrativeQA, 2WikiMQA, and Musique datasets. Regarding the hyperparameter for Twilight, we adopt their reported optimal $p$ value: $p=0.95$ for Llama3-8B and $p= 0.85$ for Mistral.
As shown in the results, \method achieves performance comparable to Twilight on both models. Notably, both methods frequently surpass the full-cache baseline across these datasets. Though, as Twilight requires different optimal $p$ values for each model, hyperparameter tuning and selection still remain necessary. Nonetheless, both \method and Twilight share the same principle of adaptive pruning. We hope our proposed objective and method could further advance the research in this direction.

\subsection{Efficiency Analysis}
\label{ssec:time}

\paragraph{End-to-End Latency}
In this section, we conduct a quantitative analysis of the latency of \method and its overall impact on inference time, beyond the memory savings from KV-cache pruning. We compare runtime on six datasets using Llama3-8B and Llama3-70B, evaluating \method against three baselines under a 50\% budget setting. Table~\ref{time} reports both the latency of the pruning operation (\textbf{Prune}) and the average generation time per sample after the prefilling stage (\textbf{Overall}).

The results show that the pruning latency of \method is comparable to prior methods. Notably, by automatically adapting to achieve higher compression ratios, \method attains the best overall generation time in 8 out of 12 comparisons, indicating a clear advantage in end-to-end generation speed. This trend remains consistent across model scales, further underscoring the efficiency aspect of \method.

\paragraph{Batched Processing}
We further conduct a comprehensive analysis of latency and throughput across varying batch sizes, as reported in Table~\ref{tab:throughput}. Following the FastGen setup, we use inputs from NarrativeQA, and perform end-to-end latency evaluations on a single A100 GPU, with the standard HuggingFace (HF) Accelerate library as the baseline. \method consistently improves throughput over naive generation by 10–20\%. Especially, it retains its performance edges robustly with increasing batch sizes.

Overall, Table~\ref{time} and Table~\ref{tab:throughput} highlight the efficiency of \method. The additional latency introduced by pruning is minimal, owing to the trivial overhead of the lightweight two-stage design. Meanwhile, the reduced KV cache lowers attention computation costs, resulting in improved inference latency and overall throughput in the end.

\subsection{Ablation Studies}
\label{ssec:ablation}

We conduct ablation studies to investigate the impact of various configurations of \method. We use Llama-3-8B-Instruct and perform experiments across five datasets, with results presented in Table~\ref{ablation}. Appendix~\ref{app:ablation} provides additional results and analysis from our ablation studies.

\begin{figure*}[t]
    \centering
    \includegraphics[width=0.85\textwidth]{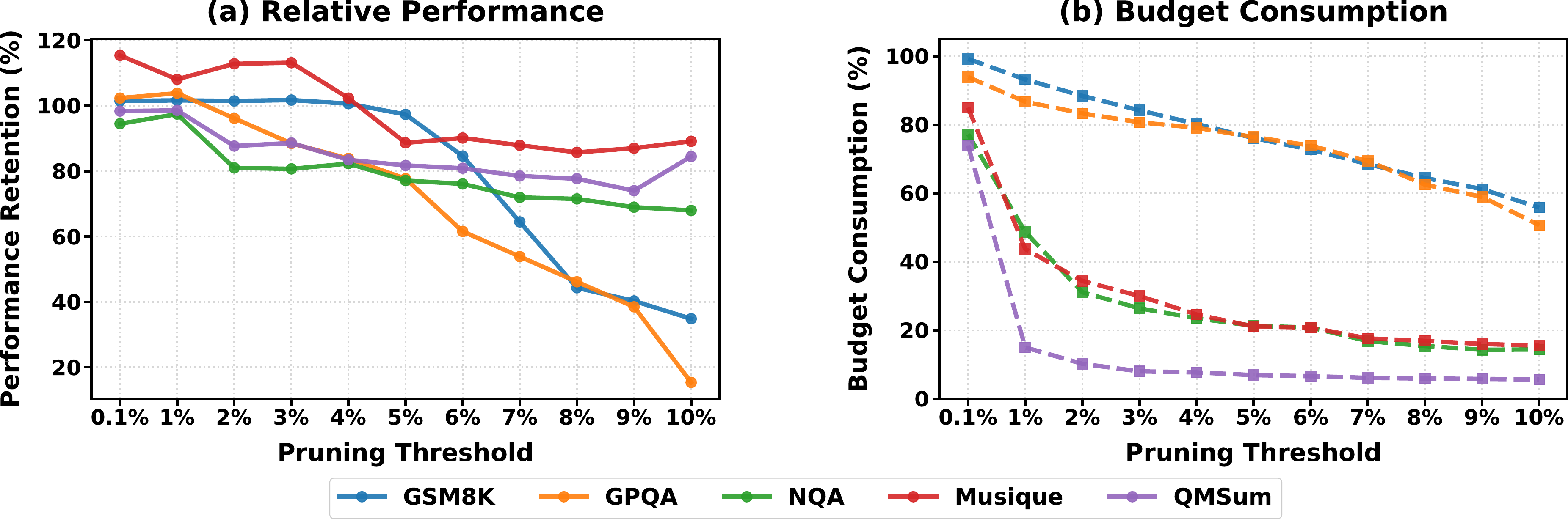}
    \caption{\textbf{Performance vs. Efficiency Trade-off.} (a) Performance retention across five datasets (solid lines). (b) Computational budget consumption (dashed lines) relative to the dense baseline. The shared legend indicates the datasets. Results show that setting the universal threshold to $1\%$ could well balance between performance and memory budget, as it maintains robust full-cache performance while substantially reducing KV cache.}
    \label{fig:combined_tradeoff}
\end{figure*}
\input{tab/abl}

\paragraph{Attention Matrix Reduction}
The reduced attention matrix $A'$ in Section~\ref{ssec:method} aggregates attention scores from the last $k$ rows. The upper part of Table~\ref{ablation} illustrates the model's performance and the actual budget when setting $k=1$, $1\%n$, $5\%n$, and $10\%n$ ($n$ being the number of input tokens). It is clear that setting $k$ as $1$ achieves a significantly reduced budget, thus a higher compression ratio, with almost no change in performance compared to $1\%n$ and $5\%n$. On the other hand, while $10\%n$ can compress more KV cache, it fails to maintain performance (for instance, on NarrativeQA, the former achieves a performance of 9.74 using 32\% of the budget, whereas the latter scores 21.44 using 48.7\% of the budget). 
What's even better is that since $k=1$ requires the least amount of computation, relying solely on the scores from the last row, the complexity of obtaining $A'$ becomes $O(1)$, independent of the sequence length. The advantages of both high efficacy and low overhead make $k=1$ a solid design choice for calculating the norm metric.

\paragraph{Initial Ranking Strategies}
Apart from the position-based ranking described in Section~\ref{ssec:method}, we also investigate other alternatives, such as ranking by each token's average attention score received from other tokens, similar to the approach in H2O \cite{DBLP:conf/nips/Zhang00CZC0TRBW23}. The results, reported at the bottom of Table~\ref{ablation}, show that across different attention-score thresholds, this attention-based ranking fails to maintain robust final performance, suggesting that raw attention scores alone are not reliable indicators of KV-cache importance.
Empirical validation supports that position ranking is an appealing choice, offering both superior efficacy and efficiency.

\paragraph{The Universal Threshold}
With the universal threshold for the attention norm difference set to 1\%, we further examine the effects of both smaller and larger values, as shown in Figure~\ref{fig:combined_tradeoff}. Intuitively, a larger threshold yields higher compression at the cost of potential performance degradation, while a smaller threshold has the opposite effect.

When the threshold is reduced to 0.1\%, the average budget increases as expected, yet model performance shows negligible improvement. In contrast, increasing the threshold to 10\% results in more aggressive KV-cache pruning but leads to substantial performance degradation. These observations suggest that 1\% provides a reasonable and robust trade-off for general use across models and tasks.

\input{tab/size}

\paragraph{Generalization}
We further evaluate whether our design and hyperparameters generalize to LLMs of different scales. As shown in Table~\ref{size}, \method applied to Llama3-70B and Qwen2.5-32B/72B consistently achieves near full-cache performance under the same configuration used in Table~\ref{main_result}. Notably, the average compression ratio improves on datasets with longer contexts, reaching close to 50\%, while maintaining comparable performance across model sizes and datasets without any threshold tuning.

%% file: tab/main.tex
\begin{table*}[!t]
\centering
\resizebox{\textwidth}{!}{
\begin{tabular}{c|ccccccccccccccc}
\toprule[1.3pt]
                & \multirow{4}{*}{~~\textbf{Methods}} & \multicolumn{3}{c}{\textbf{Math \& Science}}  & \multicolumn{2}{c}{\textbf{CR}} & \multicolumn{2}{c}{\textbf{Single-Doc QA}} & \multicolumn{2}{c}{\textbf{Multi-Doc QA}} & \multicolumn{2}{c}{\textbf{Summarization}} & \multicolumn{1}{c}{\textbf{FSL}}  & \textbf{Code} &\multirow{4}{*}{~~\textbf{Avg.}} \\  \cmidrule(lr){3-5} \cmidrule(lr){6-7}  \cmidrule(lr){8-9} \cmidrule(lr){10-11} \cmidrule(lr){12-13} \cmidrule(lr){14-14} \cmidrule(lr){15-15}             
          &                          & \rotatebox[origin=c]{45}{GSM8K}&\rotatebox[origin=c]{45}{GPQA}&\rotatebox[origin=c]{45}{TheoQA} & \rotatebox[origin=c]{45}{ThQA}  & \rotatebox[origin=c]{45}{CoQA}                  & \rotatebox[origin=c]{45}{NrtvQA}           & \rotatebox[origin=c]{45}{Qasper}            & \rotatebox[origin=c]{45}{2WkMQA}        & \rotatebox[origin=c]{45}{Musique}        & \rotatebox[origin=c]{45}{QMSum}         & \rotatebox[origin=c]{45}{M-News}                    & \rotatebox[origin=c]{45}{TriviaQA}              & \rotatebox[origin=c]{45}{Lcc} & \\ \hline
\multirow{6}{*}{\rotatebox{90}{\textbf{Llama3-8B-Instruct  }}}      & \cellcolor{blue!20}Full                      & \cellcolor{blue!20}75.28 & \cellcolor{blue!20}29.02&\cellcolor{blue!20}21.29 & \cellcolor{blue!20}25.59 
  &\cellcolor{blue!20}52.74                 & \cellcolor{blue!20}24.06              & \cellcolor{blue!20}43.91             & \cellcolor{blue!20}35.33           & \cellcolor{blue!20}14.77          & \cellcolor{blue!20}22.27         & \cellcolor{blue!20}27.37                  & \cellcolor{blue!20}70.26                 &   \cellcolor{blue!20}19.16  &\cellcolor{blue!20}100\% \\
          & \cellcolor{teal!20}Ours$_{k=1}$                     & \cellcolor{teal!20}\textbf{76.50} &\cellcolor{teal!20}\textbf{30.13} & \cellcolor{teal!20}23.03 &  \cellcolor{teal!20}\textbf{26.09}  &  \cellcolor{teal!20}\textbf{52.86}            &\cellcolor{teal!20} 23.44              & \cellcolor{teal!20}37.38             &   \cellcolor{teal!20}\textbf{36.21}              & \cellcolor{teal!20}\textbf{15.96}          & \cellcolor{teal!20}21.95         & \cellcolor{teal!20}\textbf{27.88}                    &        \cellcolor{teal!20}64.24                &  \cellcolor{teal!20}19.31  &\cellcolor{teal!20}\textbf{+0.12\%}  \\
          & \cellcolor{pink!30}\textit{Budget}              & \cellcolor{pink!30}\textit{93.2\%} & \cellcolor{pink!30}\textit{86.7\%}& \cellcolor{pink!30}\textit{92.8\%}& \cellcolor{pink!30}\textit{78.7\%}   & \cellcolor{pink!30}\textit{81.5\%}                 & \cellcolor{pink!30}\textit{48.7\%}              & \cellcolor{pink!30}\textit{46.4\%}             &      \cellcolor{pink!30}\textit{45.8\%}           & \cellcolor{pink!30}\textit{43.7\%}          & \cellcolor{pink!30}\textit{15.0\%}        & \cellcolor{pink!30}\textit{76.4\%}                   &     \cellcolor{pink!30}\textit{41.0\%}                 &   \cellcolor{pink!30}\textit{78.0\%}  & \cellcolor{pink!30}\textit{63.68}\%\\
          & H2O$_{0.9}$           & 74.60   &28.13& 21.69 & 25.41& 52.83  &    \textbf{24.55}                &     41.73              &       33.95          &         15.23       &      \textbf{22.51}         &   27.64                        &    69.77                 &   19.13        &-0.39\%    \\
          & SLM$_{0.9}$       &   72.78 &28.79& 20.75 &25.15 &  52.83            &  23.75                  &      43.63             &   32.68              &  15.86              &      22.48         &      27.21                                      & 69.97          &  \textbf{19.66} &-0.59\%   \\
          & SnapKV$_{0.9}$                   &  70.20 &27.90& 20.08  &25.38 & 52.72                      &     24.04               &\textbf{43.93}            &   33.92              &       15.56         &   22.49            &  27.65                         &     \textbf{70.58}                &   19.05  &-1.07\% \\ 
          & PyramidKV$_{0.9}$ & 75.44 & 28.79 & \textbf{24.46} &  24.72 &52.16 &23.64 &43.39 &32.10 &14.11& 22.22 &23.70& 70.48 &19.05 & -1.59\%\\
          & CAKE$_{0.9}$ & 74.23 & 27.23 &  23.16 & 22.87 & 51.18& 21.32 & 43.78&34.65&14.20& 21.19 & 22.33 & 69.22 & 19.22&-4.17\% \\
        & H2O$_{0.5}$  & 31.08 & 2.90 &16.47&17.99 & 52.37 & 23.14&43.05 & 32.79 & 15.77 & 22.79 & 26.24 &69.83 & 19.18 & -16.17\% \\
        & SLM$_{0.5}$  & 3.41  & 6.03&8.84&18.75&44.68&20.94& 37.73 & 31.20 & 15.29 & 21.80 & 25.93 & 67.79 & 19.12 & -24.73\% \\
        & SnapKV$_{0.5}$  & 12.96 &   7.81 & 16.87& 19.52 & 52.84 & 23.40 & {43.93} & 33.98 & 15.94 & 22.67 & 26.07 & 69.66 & 19.19 & -15.57\%\\ 
        & H2O$_{0.2}$ & 2.43 & 2.46 & 6.56 & 20.67 & 50.91 & 23.54 & 42.00 & 32.50 & 15.67 & 22.16 & 23.68 & 69.88 & 19.13 & -23.33\% \\
        & SLM$_{0.2}$ & 3.21 & 3.35 & 8.82  & 18.02 & 40.11 & 20.12 & 36.45 & 29.91 & 15.01 & 21.11 & 24.68 & 66.17 & 18.69 & -28.21\%\\
        & SnapKV$_{0.2}$ & 1.29 & 1.12 & 5.89 & 20.36 & 52.00 & 23.10 & 42.78 & 32.98 & 14.01 & 22.45 & 24.14 & 69.63 & 19.34 &-24.45\% \\
        & PyramidKV$_{0.2}$ & 1.36 & 1.79 & 4.82 & 20.24 & 51.71&24.31 & 41.90& 33.81 &12.56 & 21.11 & 27.01 & 68.43 & 18.19 &  -25.33\%\\
        & CAKE$_{0.2}$ & 1.67 & 4.46 & 7.93 & 21.17 &50.66 & 22.01 & 42.18 & 33.02& 14.93 & 20.15 &24.48 & 69.92 & 19.10 &  -23.47\%\\ 
          \hline \hline
\multirow{6}{*}{\rotatebox{90}{\textbf{Mistral-7B-Instruct }}} & \cellcolor{blue!20}Full                    & \cellcolor{blue!20}33.36&\cellcolor{blue!20}29.24& \cellcolor{blue!20}6.83  & \cellcolor{blue!20}20.82 & \cellcolor{blue!20}39.68                 & \cellcolor{blue!20}28.74              & \cellcolor{blue!20}37.80             & \cellcolor{blue!20}33.87           & \cellcolor{blue!20}22.88          & \cellcolor{blue!20}22.19         & \cellcolor{blue!20}22.94                     & \cellcolor{blue!20}86.87                       &  \cellcolor{blue!20}16.08&\cellcolor{blue!20}100\%    \\
          & \cellcolor{teal!20}Ours$_{k=1}$                     & \cellcolor{teal!20}\textbf{31.54}&\cellcolor{teal!20}\textbf{29.02}& \cellcolor{teal!20}6.71& \cellcolor{teal!20}\textbf{20.81} & \cellcolor{teal!20}\textbf{39.80}                 & \cellcolor{teal!20}27.20              & \cellcolor{teal!20} 38.93             &    \cellcolor{teal!20}33.46             & \cellcolor{teal!20}22.15          & \cellcolor{teal!20}\textbf{22.39}         & \cellcolor{teal!20}22.67                       &     \cellcolor{teal!20}\textbf{86.81}                    &   \cellcolor{teal!20}15.33  &\cellcolor{teal!20}\textbf{-1.50\%} \\
          & \cellcolor{pink!30}\textit{Budget}              & \cellcolor{pink!30}\textit{89.6\%} &\cellcolor{pink!30}\textit{90.5\%}&\cellcolor{pink!30}\textit{84.2\%}& \cellcolor{pink!30}\textit{92.3\%} &\cellcolor{pink!30}\textit{84.1\%}                & \cellcolor{pink!30}\textit{78.0\%}              & \cellcolor{pink!30}\textit{97.9\%}             &     \cellcolor{pink!30}\textit{86.7\%}            & \cellcolor{pink!30}\textit{74.4\%}         &\cellcolor{pink!30}\textit{84.3\%}         & \cellcolor{pink!30}\textit{89.1\%}                    &      \cellcolor{pink!30}\textit{87.2\%}          &  \cellcolor{pink!30}\textit{89.4\%}   &\cellcolor{pink!30}\textit{86.75\%} \\
          & H2O$_{0.9}$                      & 19.18  &  24.11  &   6.96 &       20.61      &   39.74  &    27.36              &  38.38                 &        35.23         &       23.24         & 22.13              &             23.41           &     86.46                       &  16.01   &-4.29\% \\
          & SLM$_{0.9}$          &        31.16     & 27.68 &   5.89       &   19.88          & 39.15  &  27.26                  &       37.66            &      35.06           &   21.94       &    22.31             &         21.73                    & 86.63   &13.67  &-4.44\%\\
          & SnapKV$_{0.9}$                   & 27.60 &  25.67   &   \textbf{7.10}     &     20.23      &  39.76 &    27.43                &    38.27               &       35.66                          &    22.99           &           22.16        &    23.31                            &   86.46        &  16.07   &-1.90\% \\ 
          & PyramidKV$_{0.9}$ & 22.74 & 26.75 & 5.35 & 20.34 & 38.43 & \textbf{27.60}  & 38.27 & \textbf{37.62} & \textbf{25.28} & 22.18 & \textbf{24.64} & 85.53 & \textbf{16.37} & -3.15\%\\ 
          & CAKE$_{0.9}$ &25.02 & 24.10&6.12& 20.21&38.82& 27.49 & \textbf{39.19} & 32.96 & 23.55 & 22.22& 24.51 &86.36 & 15.80 & -4.14\%\\
        & H2O$_{0.5}$  & 2.50   &8.26 & 5.89&20.28 & 38.53 & 27.62 & 36.97 & 34.31 & 23.17 & 22.32 & 22.60 & 86.52 & {16.45} & -14.31\%  \\
    & SLM$_{0.5}$  & 2.43   & 6.47 & 1.14 &18.27 & 32.76 & 24.91 & 33.80 & 32.73 & 20.68 & 21.48 & 18.54 & 86.45 & 13.58 &  -27.61\%  \\
    & SnapKV$_{0.5}$   & 3.03  & 8.93 & 6.83 &19.03 & 39.22 & 27.15 & 38.27 & 34.60 & 22.94 & 21.93 & 22.68 & 86.31 & 16.18 &-13.41\%\\
    & H2O$_{0.2}$ & 1.52 & 4.46 & 4.02 & 18.45 & 35.55 & 27.28 & 33.45 & 34.36 & 23.21 & 21.80 & 20.93 & 85.99 & 14.82 & -21.25\% \\
    & SLM$_{0.2}$ & 1.21 & 1.79 & 0.54 & 18.35 & 25.45 & 24.89 & 28.39 & 29.53 & 16.74 & 20.60 & 16.75 & 80.12 & 14.27 & -35.48\%\\
    & SnapKV$_{0.2}$ & 1.44 & 0.20 & 3.35 & 17.81 & 37.44 & 26.76 & 36.02 & 33.90 & 23.05 & 22.13 & 20.88 & 84.27 & 16.20 & -22.18\%\\
    & PyramidKV$_{0.2}$ & 1.14 & 0.58 & 1.40 & 19.24 & 31.20& 26.19 & 34.53 & 37.25 & 24.55 & 22.24 & 22.18 & 86.58 & 15.76 & -23.75\%\\
    & CAKE$_{0.2}$ &4.55 & 1.03 &2.09&18.21&30.07&26.25& 36.03 & 32.39 & 22.79 & 21.53 & 23.96 & 86.61 & 15.00&-24.05\% \\
          \hline
          \hline
\multirow{6}{*}{\rotatebox{90}{\textbf{Qwen2.5-7B-Instruct  }}}      &\cellcolor{blue!20} Full                      & \cellcolor{blue!20}88.02 & \cellcolor{blue!20}31.70&\cellcolor{blue!20}29.85 &\cellcolor{blue!20}24.55 &\cellcolor{blue!20}61.43                 & \cellcolor{blue!20}\textit{20.81}              & \cellcolor{blue!20}43.17             & \cellcolor{blue!20}47.15           & \cellcolor{blue!20}30.70          & \cellcolor{blue!20}23.64         & \cellcolor{blue!20}24.24                & \cellcolor{blue!20}87.64                    &   \cellcolor{blue!20} 2.44 &  \cellcolor{blue!20}100\% \\
          & \cellcolor{teal!20}Ours$_{k=1}$                     & \cellcolor{teal!20}88.02&\cellcolor{teal!20}\textbf{31.25}&\cellcolor{teal!20}\textbf{30.39} & \cellcolor{teal!20}\textbf{24.48} & \cellcolor{teal!20}60.01                 & \cellcolor{teal!20}20.66              & \cellcolor{teal!20}42.74             &      \cellcolor{teal!20}47.20           & \cellcolor{teal!20}29.56          & \cellcolor{teal!20}22.64         & \cellcolor{teal!20}24.04                        &            \cellcolor{teal!20}\textbf{87.65}              & \cellcolor{teal!20}\textbf{3.58} &\cellcolor{teal!20}\textbf{+2.63\%}    \\
          & \cellcolor{pink!30}\textit{Budget}              & \cellcolor{pink!30}\textit{90.7\%}& \cellcolor{pink!30}\textit{88.8\%} &  \cellcolor{pink!30}\textit{86.5\%}& \cellcolor{pink!30}\textit{69.2\%} &  \cellcolor{pink!30}\textit{84.1\%}                 & \cellcolor{pink!30}\textit{65.1\%}              & \cellcolor{pink!30}\textit{69.2\%}             &    \cellcolor{pink!30}\textit{73.3\%}             & \cellcolor{pink!30}\textit{64.4\%}          & \cellcolor{pink!30}\textit{70.6\%}         & \cellcolor{pink!30}\textit{85.4\%}                      &     \cellcolor{pink!30}\textit{56.6\%}             & \cellcolor{pink!30}\textit{84.4\%}    &\cellcolor{pink!30}\textit{76.02\%} \\
          & H2O$_{0.9}$                &  83.47   & 26.56 & 30.25&24.40 & \textbf{61.29}            &      20.85           &        43.26            &    \textbf{47.90}             &  \textbf{30.73}              &    23.42           &   \textbf{24.35}                             &           87.57               &   2.45  &-1.46\% \\
          & SLM$_{0.9}$                       & \textbf{88.93}  & 30.80 &28.25& 24.30 &  60.91                     &            19.86        &    42.54               &   46.02              &       28.75         &  23.06             &    23.42                         &    87.03                      &   3.05 &-0.41\%  \\
          & SnapKV$_{0.9}$                    &80.14   &30.13& 28.11   & 24.30&  61.26                    &         20.90           &    \textbf{43.31}               &   47.81              &          30.69      &      23.90         &    24.29                    &       87.57                   &    2.55  & -1.01\% \\ 
          & PyramidKV$_{0.9}$ & 84.15 & 27.68 & 26.78 & 23.51 & 50.34 & 27.60  & 42.27 & 37.62 & 25.28 & 22.18 & 21.65 & 87.53 & 3.37&-2.76\% \\
          & CAKE$_{0.9}$ &85.67 & 30.23& 29.28&23.52 & 58.27 &\textbf{29.27} & 33.73 & 46.60 & 30.32 & \textbf{23.91} & 23.89 & 90.14 & 2.41 & -0.07\%\\ 
        & H2O$_{0.5}$   &34.12 & 14.73& 20.88 &23.10 &59.69& 21.59 & {43.37}&47.60 & {30.81}  & 21.00 &22.95 & 85.54 & 2.24 & -13.47\% \\
    & SLM$_{0.5}$  &3.26&1.34&10.58 &23.75&49.87& 19.29 &36.45&42.42&25.33&21.24&22.77&87.33&3.40&-23.56\%\\
    & SnapKV$_{0.5}$  &20.77&12.28&22.76 & 22.25 & 60.21 & 20.93 & 43.22 & 47.45 & 30.61 & 23.76 & 22.70&87.57&2.21& -14.39\% \\  
    &H2O$_{0.2}$ & 5.91 & 3.57 & 17.40 & 21.70 & 55.39 & 21.41 & 40.30 & 45.84 & 29.72 & 23.04 & 20.84 & 87.99 & 2.15 & -21.77\% \\
    & SLM$_{0.2}$ & 1.06 & 4.24 & 4.82 & 23.85 & 40.92 & 18.17 & 29.22 & 39.27 & 21.71 & 20.32 & 18.94 & 77.80 & 1.46 & -37.22\%\\
    & SnapKV$_{0.2}$ & 2.58 & 7.37 & 15.13 & 21.52 & 57.06 & 21.40 & 40.93 & 46.96 & 29.87 & 23.58 & 20.69 & 87.10 & 2.44 & -20.27\%\\
    & PyramidKV$_{0.2}$ & 9.70 & 8.57 & 9.60 & 26.42 & 51.19 & 26.42 & 39.48 & 39.54 & 25.73 & 21.02 & 21.65 & 85.31 & 1.76 & -23.47\%\\
    &CAKE$_{0.2}$ &7.58& 9.11 &5.56 &23.66 &49.19& 30.20 & 43.04 & 46.35 & 29.00 & 23.33 & 22.94 & 85.45 & 2.89 & -16.98\%\\ 
    \bottomrule[1.3pt]  
\end{tabular}}
\caption{\small{Performance of \method and its comparison with five KV pruning methods on 13 datasets. \textbf{Bold} numbers indicate the best results \textbf{aside from full-cache}. \textit{Italics} represent the real budget utilized by \method. The correspondence between abbreviations and their full names of datasets can be found in Appendix~\ref{app:datasets}. \textbf{Avg.} calculates the mean ratio of the model's performance using different KV cache compression methods to its performance with the full cache. All average results (except for budget) are adjusted by subtracting 1 to provide a more intuitive understanding of the effectiveness of different methods.\label{main_result}}}
\end{table*}

%% file: tab/twi.tex
\begin{table}[!t]
\resizebox{\columnwidth}{!}{
\begin{tabular}{l|ccccc}
\toprule[1.3pt]
& \textbf{GSM8K} & \textbf{NQA} & \textbf{Qasper} & \textbf{2WQA} & \textbf{Musique} \\ \cmidrule{1-6}
Llama3-8B                  & 75.28 & 24.06       & 43.91  & 35.33    & 14.77   \\
+ Ours$_{k=1}$  & \bf 76.50 & \bf 23.44       & 37.38  & \bf 36.21    & \bf 15.96   \\
+ Twi$_{p=0.95}$ & 76.04 & 23.37       & 43.08  & 36.18    & 14.92   \\\cmidrule{1-6}
Mistral-7B &  33.36 & 28.74  & 37.80  & 33.87    & 22.88   \\ 
+ Ours$_{k=1}$  & \bf 31.54 &  \bf 27.20       &  \bf 38.93  &  \bf 33.46    &  22.15   \\
+ Twi$_{p=0.85}$ &  30.33 &  27.17       &  38.90  &  33.23    &  22.92    \\
\bottomrule[1.3pt]
\end{tabular}
}\caption{Performance comparison between \method and Twilight across five datasets using Llama3-8B-Instruct and Mistral-7B-V0.3. \textbf{Bold} numbers indicate the best results aside from full-cache.}
\label{tab:twi}
\end{table}

%% file: tab/time.tex
\begin{table*}[!t]
\centering
\resizebox{\textwidth}{!}{
\begin{tabular}{l|cccccccccccccc}
\toprule[1.3pt]
& \multicolumn{2}{c}{\textbf{GSM8K}} & \multicolumn{2}{c}{\textbf{CoQA}} & \multicolumn{2}{c}{\textbf{NarrativeQA}} & \multicolumn{2}{c}{\textbf{Musique}} & \multicolumn{2}{c}{\textbf{QMSum}} & \multicolumn{2}{c}{\textbf{TriviaQA}} & \multicolumn{2}{c}{\textbf{Avg.}} \\ \cmidrule(lr){2-3} \cmidrule(lr){4-5} \cmidrule(lr){6-7} \cmidrule(lr){8-9} \cmidrule(lr){10-11} \cmidrule(lr){12-13} \cmidrule(lr){14-15}
                                  & Overall          & Prune           & Overall              & Prune      & Overall             & Prune              & Overall           & Prune            & Overall          & Prune           & Overall                & Prune        & Overall          & Prune         \\ \midrule
\multicolumn{15}{c}{\textit{Llama3-8B-Instruct}}                                \\\arrayrulecolor{gray!60}
\midrule
Full                              & 4.693            & ---              & \textbf{0.289}       & ---         & 3.441               & ---                 & 3.659             & ---               & 5.458            & ---             & 2.717                  & ---           &      3.376            &         ---      \\
Ours$_{k=1}$                              & \textbf{4.638}   & 0.034           & 0.331                & 0.033      & \textbf{3.159}      & \textbf{0.255}     & \textbf{3.658}    & 0.264            & \textbf{5.330}   & 0.241           & 2.684                  & 0.211        &   \textbf{3.300}               &      0.173          \\
H2O$_{0.5}$                           & 4.686            & 0.020           & 0.290                & \textbf{0.018}      & 3.243               & 0.264              & 3.679             & 0.266            & 5.398            & 0.249           & \textbf{2.612}         & 0.237        & 3.318                 &  0.176             \\
SLM$_{0.5}$                           & 6.071            & \textbf{0.006}  & 0.301                & 0.030      & 3.230               & 0.257              & 3.784             & \textbf{0.259}   & 5.454            & \textbf{0.234}           & 2.628                  & \textbf{0.208}        &   3.578            &     \textbf{0.166}          \\
SnapKV$_{0.5}$                        & 5.874            & 0.021           & 0.285                & 0.019      & 3.307               & 0.265              & 3.721             & 0.266            & 5.519            & 0.264           & 2.650                  & 0.265        &          3.559        &              0.183
\\\arrayrulecolor{black}
\midrule
\multicolumn{15}{c}{\textit{Llama3-70B-Instruct}}                                          \\\arrayrulecolor{gray!60}
\midrule
Full                              & 17.504           &    ---             & 1.167                &     ---       & 6.285               &        ---            & \textbf{6.565}    &   ---               & 13.993           &      ---           & 5.302                  &    ---          &      8.469            &        ---       \\
Ours$_{k=1}$                              & \textbf{15.975}  & 0.154           & 1.253                & 0.290      & \textbf{6.042}      & 2.345              & 7.412             & 2.352            & \textbf{13.835}  & 3.028           & \textbf{5.224}         & 1.602        & \textbf{8.290}                 &    1.623          \\
H2O$_{0.5}$                           & 19.062           & \textbf{0.114}  & \textbf{1.059}                & 0.240      & 6.788               & 2.298              & 7.082             & 2.307            & 14.360           & 3.712           & 5.566                  & 1.308        &      8.986            &  1.663             \\
SLM$_{0.5}$                           & 22.079           & 0.150           & 1.138                &\textbf{0.229}     & 6.804               & \textbf{2.069}              & 7.107             & \textbf{2.053}            & 14.734           & 3.700           & 5.576                  & 1.367        &         9.573         &    1.595           \\
SnapKV$_{0.5}$                        & 16.517           & 0.117           & 1.116                & 0.241      & 6.795               & 2.474              & 7.085             & 2.468            & 13.846           & \textbf{2.793}  & 5.562                  & \textbf{1.083 }       &      8.487            &   \textbf{1.529}           \\ 
\arrayrulecolor{black}
\bottomrule[1.3pt]
\end{tabular}
}
\caption{The average inference time and pruning time of Llama3 with size of 8B/70B across six datasets, measured in seconds, with lower values indicating better performance.\label{time}}
\end{table*}

%% file: tab/abl.tex
\begin{table}[!t]
\centering
\resizebox{\columnwidth}{!}{
\begin{tabular}{cccccc}
\toprule[1.3pt]
& \textbf{GSM8K} & \textbf{CoQA}  & \textbf{NQA} & \textbf{Musique} & \textbf{QMSum}   \\ \cmidrule{1-6}
Full      & 75.28 & 52.74 & 24.06  & 14.77   & 22.27 \\
\midrule
\midrule
Pos$_{t=1\%}$ & \multicolumn{5}{c}{\textit{Performance Using Different $k$}} \\
\midrule
$k=1$   &  76.50     & 52.86      & 23.44       & 15.96        &    21.95 \\
\textit{Budget}    &  \textit{93.2\%}     &  \textit{81.5\%}     &  \textit{48.7\%}      & \textit{43.7\%}        &   \textit{15.0\%}  \\
$k=1\% n$  &  76.19     & 52.87      &   23.17     &    15.40     &   21.94    \\
\textit{Budget}    &    \textit{95.1\%}   &    \textit{94.8\%}   &   \textit{77.3\%}    &  \textit{77.2\%}       &  \textit{59.5\%}    \\
$k=5\%n$  &    75.59   &   52.75    &   21.62     &   13.71      &    21.43   \\
\textit{Budget}    &   \textit{98.6\%}   &   \textit{96.2\%}    &    \textit{45.1\%}    &  \textit{30.0\%}       &     \textit{33.1\%}  \\
$k=10\%n$   &  76.72     &  52.85     &     9.74   &  13.67       &   21.11    \\
\textit{Budget}    &  \textit{96.3\%}     &  \textit{96.8\%}    &   \textit{32.0\%}     &   \textit{19.7\%}      &   29.3\%    \\
\midrule
\midrule
$k=1$ & \multicolumn{5}{c}{\textit{Performance with Different Ranking Method}}   
\\
\midrule
Attn$_{t=1\%}$ &  2.35     &   43.68    &   13.37     & 9.58        &  17.62     \\
\textit{Budget}    &  \textit{20.0\%}     &  \textit{9.7\%}     &   \textit{6.4\%}     &    \textit{6.4\%}     &   \textit{6.4\%}    \\
Attn$_{t=0.01\%}$ &  54.66    &    51.58   &  17.45      &  14.79       &  20.90     \\
\textit{Budget}    &  \textit{61.7\%}     &  \textit{28.6\%}     &   \textit{6.7\%}     &  \textit{7.9\%}       &  \textit{7.1\%}     \\ 
\arrayrulecolor{black}\bottomrule[1.3pt]
\end{tabular}
}
\caption{Performance comparison for ablation studies in Section~\ref{ssec:ablation}: different $k$ values for attention matrix reduction, and using an alternative initial ranking strategy by attention scores with different thresholds.}
\label{ablation}
\vspace{-0.5em}
\end{table}

%% file: tab/size.tex
\begin{table}[!t]
\centering
\resizebox{\columnwidth}{!}{
\begin{tabular}{l|ccccc}
\toprule[1.3pt]
& \textbf{GSM8K} & \textbf{CoQA}  & \textbf{NQA}  & \textbf{Musique} & \textbf{QMSum}  \\
\cmidrule{1-6}
\multicolumn{6}{c}{\textit{Llama3-70B-Instruct}}                                    \\\arrayrulecolor{gray!60}\hline 
Full   & 89.69 & 60.36 & 27.15        & 29.31   & 22.52      \\
Ours   &  89.76     &   60.38    &  26.94           &   28.88    &              22.30             \\
\textit{Budget} &  \textit{93.3\%}     &  \textit{77.2\%}     &  \textit{58.7\%}           &    \textit{60.8\%}    &                 \textit{52.9\%}          \\\arrayrulecolor{black}\cmidrule{1-6}
\multicolumn{6}{c}{\textit{Qwen2.5-32B-Instruct}}   
\\\arrayrulecolor{gray!60}\hline 
Full   &  91.36     &  58.03    &   24.78         &   40.04   &    22.84                       \\
Ours   &  91.81    &  57.29     &   22.65         &  40.54      &     22.52                      \\
\textit{Budget} &  \textit{91.6\%}     &   \textit{85.0\%}    &   \textit{68.2\%}          &   \textit{74.4\%}     &    \textit{76.8\%}                       \\\arrayrulecolor{black}\cmidrule{1-6}
\multicolumn{6}{c}{\textit{Qwen2.5-72B-Instruct}}  
\\\arrayrulecolor{gray!60}\hline 
Full   &   90.22    &  54.14      &  24.36        &    42.13      &   23.93                        \\
Ours   &    90.30   &  54.19     & 24.10  &   41.70             &     23.31                      \\
\textit{Budget} &   \textit{90.9\%}    & \textit{87.6\%}      &  \textit{63.4\%} &  \textit{65.0\%}          &     \textit{68.6\%}      \\ \arrayrulecolor{black}\bottomrule[1.3pt]               
\end{tabular}}
\caption{Performance of different LLMs with scales of 70B, 32B, and 72B across five datasets with the exact same \method configuration as in Table~\ref{main_result}, demonstrating the generalization of \method.\label{size}}
\end{table}

%% file: sections/conclusion.tex
\section{Conclusion}

In this study, we introduce a new KV cache compression objective that lifts the threshold dependency, so to achieve input-insensitive pruning for robust inference performance.
Towards this objective, we propose a novel method, termed \method, which employs a straightforward yet effective two-stage KV cache pruning process.
Comprehensive experiments conducted across diverse datasets, encompassing a variety of tasks, context lengths and LLM models, demonstrate that \method achieves near-lossless compression robustly without involving any input-specific thresholds, while able to deliver notable KV cache reduction.

%% file: apdx/freeze_old.tex
\section{Full Algorithm of \method}

\begin{algorithm}
\caption{\method \label{al}}
\begin{algorithmic}
    \STATE \textbf{Input:} Prompt, Threshold $t$
    \STATE \textbf{Output:} Compressed KV Cache
    \STATE Create Empty List $K_c,V_c$
    \FOR{\textit{Transformer Layer $L_i$ in LLM}}
        \STATE $Q^i,K^i,V^i \leftarrow L_i(\text{Prompt})$
        \STATE $R^i \gets$ Postion-Based Importance Rank
        \STATE $A_{last}^i\gets \operatorname{Attention}(Q^i[\dots,-1,:], K^{iT})$
        \STATE $F_{b}^i \gets \operatorname{Frobenius}(A_{last}^i)$
        \STATE $A_{last}^i\gets \operatorname{Square}(A_{last}^i)$
        \STATE Reorder $A_{last}^i$ by Rank $R$
        \STATE $A_{cumsum}^i \gets \operatorname{Cumsum}(A_{last}^i)$
        \STATE $A_{cumsum}^i \gets \operatorname{Sqrt}(A_{cumsum}^i)$
        \STATE $A_{ratio}^i \gets (F_b^i - A_{cumsum}^i)/F_b^i$
        \STATE $\text{Index}^i \gets \operatorname{Max}(\operatorname{Where}(A_{ratio}^i<=t))$
        \STATE $K_c^i \gets \operatorname{Compress} K^i \text{by} R^i[I:]$
        \STATE $V_c^i \gets \operatorname{Compress} V^i \text{by} R^i[I:]$ 
        \STATE Append $K_c^i,V_c^i$ to $K_c,V_c$
    \ENDFOR
    \RETURN $K_c,V_c$
\end{algorithmic}
\end{algorithm}

\section{Retain Bottom LLM Layers}
\label{app:freeze}
As described in Section~\ref{ssec:method}, the KV-cache pruning in \method begins from the third LLM layer, which also aligns with design choice in related methods, e.g., \textsc{Scissorhands} \cite{DBLP:conf/nips/LiuDLWXXKS23} that allocates less budget in early
layers.

In this part, we empirically justify this design choice. We apply \method with Llama3-8B and evaluate it on GSM8K and CoQA. We first present a case study, followed by a comparison examining the impact of not retaining/freezing the full KV cache in the first two layers.

As shown in Table~\ref{case}, when the threshold is set to 1\% and no layers are frozen, the outputs for two examples on GSM8K and CoQA are both incorrect and lack logical coherence, despite the overall budget exceeding 90\% in each case. Examining the per-layer budgets reveals that layers 1 and 2 have relatively low budgets, whereas layers 3 through 31 exhibit uniformly high budgets, with the final layer dropping again.
We hypothesize that, in the first two layers, the model has not yet formed sufficiently discriminative attention patterns to identify truly important tokens, leading to a relatively uniform attention distribution. This can result in the premature eviction of important tokens, preventing later layers from accessing critical information and ultimately causing generation failures.

Based on the above case study, we further explore retaining specific layers to identify an optimal configuration. We evaluate different layer-freezing strategies, with partial results summarized in Table~\ref{freeze}. We observe that freezing the first two layers can strike a balance between model performance and budget. In contrast, additionally retaining the 31st layer yields negligible performance gains while incurring a higher budget. Accordingly, \method applies KV-cache compression starting from the third layer.

\begin{table}[!t]
\centering
\resizebox{\columnwidth}{!}{
\begin{tabular}{ll}
\toprule[1.3pt]
\multicolumn{2}{c}{\textbf{GSM8K}}             \\ \cmidrule{1-2}
Input            & \begin{tabular}[c]{@{}l@{}}A robe takes 2 bolts of blue fiber a-\\nd half that much white fiber. How\\many bolts in total does it take?\end{tabular}             \\ \cmidrule{1-2}
Budget & \begin{tabular}[c]{@{}l@{}}\textit{Layer 1}: \textcolor{red}{{67.71}}\\ \textit{Layer 2}: \textcolor{red}{{82.29}}\\ \textit{Layer 3$\sim$31:} 95.83\\ \textit{Layer 32}: 37.50\\\textbf{Avg.}: 93.90\end{tabular}   \\ \cmidrule{1-2}
Output          & \begin{tabular}[c]{@{}l@{}}I have determined by answering 
the\\answer to the format of bolts bolts b-\\olts...(repeat) \end{tabular}                            \\  \cmidrule{1-2}
Ground-Truth     & 3                  \\\toprule[1.3pt]
\multicolumn{2}{c}{\textbf{CoQA}}              \\  \cmidrule{1-2}
Input            & \begin{tabular}[c]{@{}l@{}}You are given a story and a question.\\Answer the question as concisely as\\you can...Question: What color was\\Cotton?\end{tabular} \\ \cmidrule{1-2}
Budget & \begin{tabular}[c]{@{}l@{}}\textit{Layer 1}: \textcolor{red}{{42.14}}\\ \textit{Layer 2}: \textcolor{red}{{42.54}}\\ \textit{Layer 3$\sim$31}: 99.19\\ \textit{Layer 32}: 79.64\\\textbf{Avg.}: 95.03\end{tabular}   \\ \cmidrule{1-2}
Output           & \begin{tabular}[c]{@{}l@{}}Question: Question: What is the que-\\stion: What is the question:\end{tabular}            \\ \cmidrule{1-2}
Groud-Truth      & White        \\             \bottomrule[1.3pt]                                                                                           
\end{tabular}
}
\caption{Case Study (Appendix~\ref{app:freeze}).\label{case}}
\end{table}

\begin{table}[!t]
\centering
\resizebox{\columnwidth}{!}{
\begin{tabular}{ll|cccccc}
\toprule[1.3pt]
\textbf{Datasets} && \textbf{None}  & \textbf{0}      & \textbf{0,1}    & \textbf{0,1,2} & \textbf{0,1,2,3} & \textbf{0,1,31} \\ \cmidrule{1-8}
GSM8K   & Score         & 3.9 & 20.32  & 76.50  & 76.65 & 76.35   & 76.57  \\
        & \textit{Budget}       & \textit{86.7\%}  & \textit{87.0\%} & \textit{93.2\%} & \textit{94.1\%}  & \textit{94.7\%}    & \textit{96.6\%}   \\  \cmidrule{1-8}
CoQA    & Score          & 20.89  & 51.82  & 52.86  &51.58 & 53.32   & 52.46  \\
        & \textit{Budget}       & \textit{75.5\%} & \textit{77.8\%}  & \textit{81.5\%}   & \textit{82.8\%} & \textit{83.0\%}    & \textit{82.5\%}  \\ \bottomrule[1.3pt]
\end{tabular}
}
\caption{Performance of Llama3-8B-Instruct on GSM8K and CoQA using \method varying frozen layer configurations. The column headers indicate which layers are frozen/retained during inference.\label{freeze}}
\end{table}

%% file: apdx/datasets.tex
\section{Datasets Used in Experiments}
\label{app:datasets}

\input{tab/slmf}

\begin{table}[!t]
\centering
\resizebox{\columnwidth}{!}{
\begin{tabular}{l|ccccc}
\toprule[1.3pt]
\textbf{Methods}    & \textbf{GSM8K} & \textbf{CoQA}  & \textbf{NQA} & \textbf{Musique} & \textbf{QMSum}   \\ \cmidrule{1-6}
Full      & 75.28 & 52.74 & 24.06  & 14.77   & 22.27 \\ \cmidrule{1-6}
Pos$_{t=1\%}$ & \multicolumn{5}{c}{\textit{Performance Using Different $k$}} \\\arrayrulecolor{gray!60}
\midrule
$k=1$   &  76.50     & 52.86      & 23.44       & 15.96        &    21.95 \\
\textit{Budget}    &  \textit{93.2\%}     &  \textit{81.5\%}     &  \textit{48.7\%}      & \textit{43.7\%}        &   \textit{15.0\%}  \\
$k=1\% n$  &  76.19     & 52.87      &   23.17     &    15.40     &   21.94    \\
\textit{Budget}    &    \textit{95.1\%}   &    \textit{94.8\%}   &   \textit{77.3\%}    &  \textit{77.2\%}       &  \textit{59.5\%}    \\
$k=2\% n$  &  76.19     & 52.82      &   22.87     &    14.34     &   21.74    \\
\textit{Budget}    &    \textit{96.0\%}   &    \textit{96.1\%}   &   \textit{69.8\%}    &  \textit{62.0\%}       &  \textit{47.1\%}    \\
$k=3\% n$  &  75.82     & 52.80      &   23.03     &    13.61     &   21.56    \\
\textit{Budget}    &    \textit{98.6\%}   &    \textit{95.7\%}   &   \textit{60.3\%}    &  \textit{46.0\%}       &  \textit{39.5\%}    \\
$k=4\% n$  &  75.21     & 52.76      &   22.96     &    13.45     &   21.47    \\
\textit{Budget}    &    \textit{98.8\%}   &    \textit{95.8\%}   &   \textit{51.9\%}    &  \textit{35.2\%}       &  \textit{35.4\%}    \\

$k=5\%n$  &    75.59   &   52.75    &   21.62     &   13.71      &    21.43   \\
\textit{Budget}    &   \textit{98.6\%}   &   \textit{96.2\%}    &    \textit{45.1\%}    &  \textit{30.0\%}       &     \textit{33.1\%}  \\
$k=6\% n$  &  75.66     & 52.81      &   21.95     &    14.33       &   20.94    \\
\textit{Budget}    &    \textit{98.5\%}   &    \textit{96.6\%}   &   \textit{40.4\%}    &  \textit{25.8\%}       &  \textit{31.6\%}    \\
$k=7\% n$  &  76.57     & 52.88      &   21.18     &    13.66     &   20.94    \\
\textit{Budget}    &    \textit{98.0\%}   &    \textit{96.7\%}   &   \textit{37.3\%}    &  \textit{22.5\%}       &  \textit{30.6\%}    \\
$k=8\% n$  &  76.35     & 52.86      &   20.36     &    13.75     &   21.11    \\
\textit{Budget}    &    \textit{97.4\%}   &    \textit{96.7\%}   &   \textit{35.1\%}    &  \textit{20.9\%}       &  \textit{30.1\%}    \\
$k=9\% n$  &  76.65     & 52.84      &   19.99     &    13.66     &   20.76    \\
\textit{Budget}    &    \textit{96.8\%}   &    \textit{96.8\%}   &   \textit{33.4\%}    &  \textit{19.9\%}       &  \textit{29.5\%}    \\
$k=10\%n$   &  76.72     &  52.85     &     9.74   &  13.67       &   21.11    \\
\textit{Budget}    &  \textit{96.3\%}     &  \textit{96.8\%}    &   \textit{32.0\%}     &   \textit{19.7\%}      &   29.3\%    \\
\arrayrulecolor{black}\bottomrule[1.3pt]
\end{tabular}
}
\caption{Performance comparison with different $k$ for reducing the full attention matrix, expanded from Table~\ref{ablation}.\label{extend_ablation}}
\end{table}

\begin{table*}[!]
\centering

\label{tab:performance_original_layout}
\resizebox{\textwidth}{!}{
\begin{tabular}{lcccccccccc}
\toprule
& \multicolumn{1}{c}{\textbf{BS=1}} & \multicolumn{2}{c}{\textbf{BS=2}} & \multicolumn{2}{c}{\textbf{BS=8}} & \multicolumn{3}{c}{\textbf{BS=16}} \\
\cmidrule(lr){2-2} \cmidrule(lr){3-4} \cmidrule(lr){5-6} \cmidrule(lr){7-9}

\textbf{[Input, Output]} & [4K, 8K] & [8K, 16K] & [512, 4K] & [4K, 8K] & [512,512] & [4K, 4K] & [512, 512] & [2K, 2K] \\
\hline
\multicolumn{9}{c}{\cellcolor{gray!40}\textit{Latency (s/100tokens)}} \\
HF Accelerate & 3.61 & 5.29 & 3.50 & 5.26 & 3.48 & 12.23 & 4.71 & 13.01 \\
Ours & 3.19 & 4.08 & 2.84 & 3.60 & 3.06 & 10.42 & 4.25 & 10.64 \\
\hline
\multicolumn{9}{c}{\cellcolor{gray!40}\textit{Throughput (token/s)}} \\
HF Accelerate & 27.71 & 18.89 & 57.08 & 38.01 & 230.00 & 65.41 & 339.43 & 122.98 \\
\method & 31.35 & 24.51 & 70.30 & 55.56 & 261.83 & 76.81 & 376.89 & 150.40 \\
\hline
Budget & 73.3\% & 55.9\% & 62.0\% & 34.0\% & 76.7\% & 65.3\% & 82.8\% & 78.2\% \\
\bottomrule
\end{tabular}
}
\caption{Performance comparison across various batch sizes and sequence lengths. We report latency (second/100 tokens, lower is better) and throughput (tokens/second, higher is better). More discussions regarding method efficiency are addressed in Section~\ref{ssec:time}.\label{tab:throughput}}
\end{table*}
In this section, we provide a comprehensive overview of all the tasks and datasets utilized in the experiments of this paper.
\paragraph{Math \& Science} This task evaluates the model's ability to tackle mathematical and scientific problems. By directly inputting questions and comparing the model's output with the correct answers, we calculate the model's \textit{Accuracy} on these datasets: \textbf{GSM8K} is a dataset for evaluating model's math-solving skills, featuring 8,000 elementary-level math word problems requiring basic arithmetic and reasoning. \textbf{GPQA} tests model's understanding of physics concepts and problem-solving across various topics, assessing scientific reasoning abilities. \textbf{TheoremQA} evaluates model's grasp and application of mathematical theorems, ranging from simple applications to complex proofs, testing advanced math skills.
\paragraph{Commonsense Reasoning (CR)} This task evaluates model's ability to make deductions and understand everyday situations using implicit knowledge and logical inference. \textbf{TruthfulQA} (ThQA) evaluates model's ability to generate accurate and truthful responses, testing models on distinguishing fact from fiction, especially in areas prone to misconceptions. We use \textit{BLEU} as the metric. \textbf{CoQA} assesses model's ability to understand and respond to questions in a conversational context, focusing on maintaining coherence and context throughout a dialogue. We use \textit{F1 Score} as the metric.

\paragraph{Single Document QA (Single-Doc QA)} This task assesses the model's reading comprehension skills when dealing with a single, extended document. \textbf{NarrativeQA}~\cite{DBLP:journals/tacl/KociskySBDHMG18} is a dataset designed to evaluate model's ability to comprehend and answer questions based on narrative texts, focusing on understanding stories and their underlying themes. \textbf{Qasper}~\cite{DBLP:conf/naacl/DasigiLBCSG21} is a dataset aimed at assessing model's capability to extract and answer questions from academic papers, emphasizing understanding complex scientific information. We employ \textit{F1 Score} as the metric for above two datasets.

\paragraph{Multi-Document QA (Multi-Doc QA)} This task evaluates the model's reading comprehension capabilities across multiple extended documents. \textbf{2WikiMultiHopQA} (2WKMQA)~\cite{DBLP:conf/coling/HoNSA20} is a dataset designed to test model's ability to perform multi-hop reasoning and answer complex questions using information from multiple Wikipedia articles. \textbf{MuSiQue}~\cite{DBLP:journals/tacl/TrivediBKS22} evaluates model's skill in integrating and reasoning over information from multiple sources to answer comprehensive questions accurately. We leverage \textit{F1 Score} as the metric for above two datasets.

\paragraph{Summarization} This task examines the model's ability to comprehend and summarize lengthy documents. \textbf{QMSum}~\cite{DBLP:conf/naacl/ZhongYYZMJACLQR21} is a dataset for evaluating model's ability to generate concise summaries of meeting transcripts, focusing on capturing the key points from multi-party discussions. \textbf{Multi-News} (M-News)~\cite{DBLP:conf/acl/FabbriLSLR19} is a dataset that challenges models to create coherent summaries by synthesizing information from multiple news articles on the same topic. We use \textit{Rouge-L} as the metric for above two datasets.

\paragraph{Few-Shot Learning (FSL)} This task assesses the model's few-shot learning capabilities. \textbf{TriviaQA}~\cite{DBLP:conf/acl/JoshiCWZ17} is a dataset designed to assess model's ability to retrieve and answer questions based on large collections of trivia, emphasizing comprehension and factual recall. We use \textit{F1 Score} as the metric.

\paragraph{Code} This task evaluates the model's ability to complete and generate code. \textbf{LCC}~\cite{DBLP:conf/icml/GuoXD0M23} is a dataset focused on evaluating models' ability to understand and generate code by considering extended code contexts, enhancing the ability to reason over complex programming structures. We use \textit{Edit Sim} as the metric.

%% file: tab/slmf.tex
\begin{table*}[!t]
\centering
\resizebox{0.7\textwidth}{!}{
\begin{tabular}{l|cccccc}
\toprule[1.3pt]
          & \textbf{GSM8K} & \textbf{GPQA} & \textbf{CoQA} & \textbf{NrtvQA}  & \textbf{QMSum} & \textbf{TriviaQA} \\ \cmidrule{1-7}
Full      & 75.28          & 29.02         & 52.74         & 24.06                      & 22.27          & 70.26             \\
\midrule
Ours$_{k=1}$   & 76.50          & 30.13         & 52.86         & 23.44                    & 21.95          & 64.24             \\
\textit{Budget}    & \textit{93.2\%}         & \textit{86.7\% }       & \textit{81.5\%}        & \textit{48.7\%}                   & \textit{15.0\%}         & \textit{41.0\%}           \\
\midrule
SLM$_{0.9}$    & 72.78          & 28.79         & 52.83         & 23.75                      & 22.48          & 69.97             \\
SLM$_{F0.9}$ & 76.42          & 29.24         & 52.89         & 22.95                      & 22.47          & 69.10             \\
SLM$_{0.5}$    & 3.41           & 6.03          & 44.68         & 20.94                    & 21.80          & 67.79             \\
SLM$_{F0.5}$ & 4.55           & 3.13          & 44.59         & 21.31                     & 21.88          & 68.04             \\
SLM$_{0.2}$    & 3.21           & 3.35          & 40.11         & 20.12                     & 21.11          & 66.17             \\
SLM$_{F0.2}$ & 1.21           & 1.12          & 38.44         & 19.52                      & 20.61          & 65.69     \\ \bottomrule[1.3pt]        
\end{tabular}
}
\caption{Performance comparison between \method and StreamingLLM (\textbf{SLM}) with the first two layers frozen (\textbf{SLM$_F$}) under difference KV cache budgets ($0.9$, $0.5$ and $0.2$).}\label{slm_f}
\end{table*}

%% file: apdx/ablation.tex
\section{Ablation}
\label{app:ablation}

In this section, we present additional ablation study results for Section~\ref{ssec:ablation}. By setting various values for $k$ in reducing the full attention matrix, we expand upon the results shown in Table~\ref{ablation}, which are shown in Table~\ref{extend_ablation}. These experiments facilitate a deeper understanding of how different parameter settings impact model performance and provide a basis for optimizing parameter selection.

As shown in Table~\ref{extend_ablation}, setting $k=1$ not only conserves pruning time but also achieves better model performance with a reduced budget, making it an ideal design choice.

%% file: apdx/slmf.tex
\section{Comparison with StreamingLLM}
\label{app:slmf}

We conduct additional experiments on six datasets using StreamingLLM (with Llama3-8B) while freezing the first two layers, as an ablation to assess the impact of the proposed two-stage process. As shown in Table~\ref{slm_f}, despite adopting the same layer-freezing strategy as \method, StreamingLLM fails to achieve the objective of this work, as it cannot consistently maintain optimal performance across datasets such as GSM8K and GPQA under varying budgets. These results indicate that relying solely on the \emph{position bias}, as in StreamingLLM, is insufficient to preserve full performance, and that incorporating additional design signals is necessary for achieving dynamic, lossless pruning.